\pdfoutput=1

\documentclass[journal]{IEEEtran}
\usepackage{mflogo,texnames}
\usepackage{enumerate}
\usepackage{float}
\usepackage{hyperref}       
\usepackage{url}            
\usepackage{booktabs}       
\usepackage{amsfonts}       
\usepackage{nicefrac}       
\usepackage{microtype}      
\usepackage{times}
\usepackage{xcolor}
\usepackage{soul}

\usepackage{lineno,hyperref}
\usepackage{enumerate}
\usepackage{amsfonts}       
\usepackage{graphicx}  
\usepackage{subfigure}
\usepackage{multirow}
\usepackage{amsmath}
\usepackage{float}
\usepackage{color}
\usepackage[ruled]{algorithm2e}

\ifCLASSINFOpdf
\else
\fi
\begin{document}
	%
	\title{Dyn-Backdoor: Backdoor Attack on Dynamic\\ Link Prediction}
	%
	%
	%
	
	\author{Jinyin Chen,
		Haiyang Xiong,
		Haibin Zheng,
		Jian Zhang,
		Guodong Jiang
		and Yi Liu
		\thanks{J. Chen is with the Institute of Cyberspace Security, Zhejiang University
			of Technology, Hangzhou, 310023, China.}
		\thanks{J. Chen, H. Xiong, H. Zheng, Jian Zhang, G. Jiang are with the College of Information Engineering,
			Zhejiang University of Technology, Hangzhou 310023, China.}
		\thanks{Y. Liu is with the Institute of Process Equipment and Control Engineering, Zhejiang University of Technology, Hangzhou, 310023, China.}
		\thanks{Manuscript received xxxx xx, xxxx; revised xx xx, xxxx. This research was supported by the National Natural Science
			Foundation of China under Grant No. 62072406, the
			Natural Science Foundation of Zhejiang Province under Grant
			No. LY19F020025.
	}}
	
	%
	%

	\markboth{Journal of \LaTeX\ Class Files,~Vol.~14, No.~8, August~2015}%
	{Shell \MakeLowercase{\textit{et al.}}: Bare Demo of IEEEtran.cls for IEEE Journals}
	%



	\maketitle
	
	\begin{abstract}
		Dynamic link prediction (DLP) makes graph prediction based on historical information. Since most DLP methods are highly dependent on the training data to achieve satisfying prediction performance, the quality of the training data is crucial. Backdoor attacks induce the DLP methods to make wrong prediction by the malicious training data, i.e., generating a subgraph sequence as the trigger and embedding it to the training data. However, the vulnerability of DLP toward backdoor attacks has not been studied yet. To address the issue, we propose a novel backdoor attack framework on DLP, denoted as Dyn-Backdoor. Specifically, Dyn-Backdoor generates diverse initial-triggers by a generative adversarial network (GAN). Then partial links of the initial-triggers are selected to form a trigger set, according to the gradient information of the attack discriminator in the GAN, so as to reduce the size of triggers and improve the concealment of the attack. Experimental results show that Dyn-Backdoor launches successful backdoor attacks on the state-of-the-art DLP models with success rate more than 90\%. Additionally, we conduct a possible defense against Dyn-Backdoor to testify its resistance in defensive settings, highlighting the needs of defenses for backdoor attacks on DLP.

	\end{abstract}
	

	%
	\IEEEpeerreviewmaketitle

	\section{Introduction}
	%
	%
	%
	%
	\IEEEPARstart
	{G}{raphs} are usually used to describe complex systems \cite{20On}, \cite{17A_fea}, \cite{19Detecting} in various fields, such as social networks \cite{18Belief}, \cite{20publishing}, biological networks \cite{1992Prediction}, electric systems \cite{Gao2012A}, economics \cite{2015Correlation}, neural networks \cite{2015Stochastic}, etc. In practice, most of these systems are evolving with time, which can be modeled as dynamic networks with nodes or links appearing and disappearing over time \cite{2017Complex}, \cite{2018Optimizing}. For instance, the purchases among users and items in e-commerce platforms, e.g. Taobao, Amazon, are updated in real-time. To predict the next time 
	purchase behavior in the e-commerce platform can be defined as a link prediction on dynamic networks. Specifically, dynamic link prediction (DLP) tries to predict the existence of links at a certain time in the future based on the historical information of the network, which helps to better understand the evolving pattern of the network \cite{2015Link}. Taking the e-commerce platform as an example, it will make targeted product recommendation benefiting from the DLP on basis of the users' historical purchase records.
	
	Numerous DLP methods have been proposed. For example, Yao et al. \cite{2016Link} redefined the indicators for link prediction to extend them from static networks to dynamic ones, i.e., redefining common neighbor based on timestamps with different weights. Zhang et al. \cite{2017Efficient} used the improved resource allocation for DLP by recalculating the updated part of the graph rather than the entire graph. For random walk based approaches \cite{2010Random}, \cite{Liu2016Sampling}, \cite{Ahmed2016An}, \cite{18Continuous-Time}, they can reduce the complexity of the model, since they usually take a walk of the local structure. 
	With the success of deep learning applied to computer vision, DLP methods based on deep learning \cite{DBLP:journals/corr/abs-1805-11273}, \cite{2019dyngraph2vec}, \cite{2018Evolving}, \cite{2018Deep}, \cite{2020EvolveGCN}, \cite{2019Generative} are thoroughly studied. They mainly use the nonlinear and hierarchical nature of neural networks to  capture the evolving pattern of dynamic networks. In particular, the experiments in research \cite{2018Deep}, \cite{0E} demonstrate that DLP methods based on deep learning  generally outperform the non-deep ones.
	
	\begin{figure*}[]
		\centering
		\includegraphics[width=\linewidth]{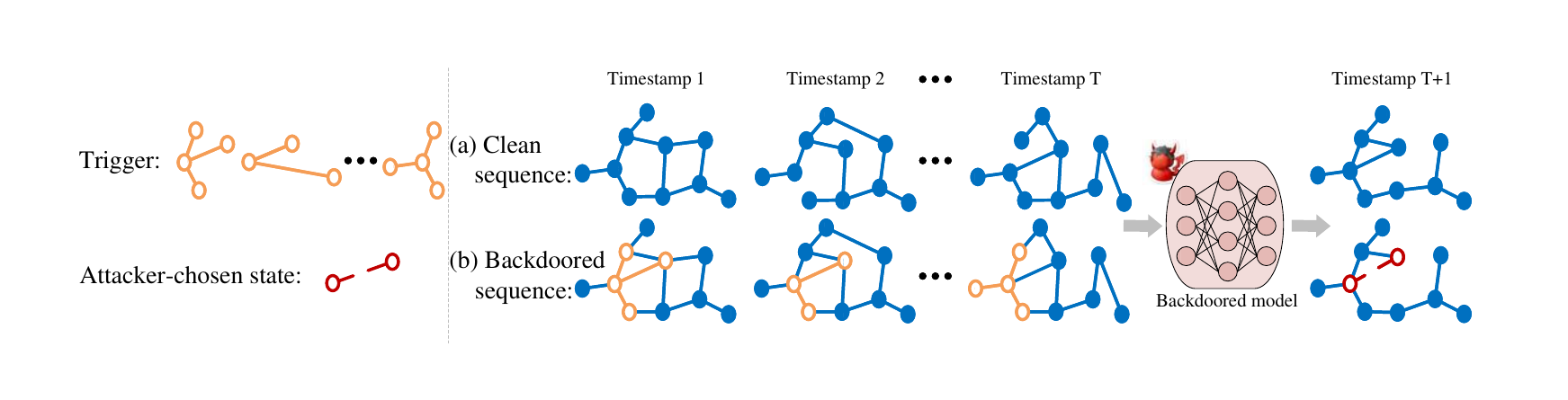}
		\centering
		\setlength{\abovecaptionskip}{-1.cm}
		\caption{An illustration of backdoor attack on DLP. On the left side, the orange subgraph sequence is defined as the backdoor trigger for training and testing, and the red link represents the target link selected by the attacker. Specifically, the existence links are represented by solid lines, while the non-existent links are represented by dashed lines or wireless ones. The right side shows the behavior of an already backdoored model when fed by a clean sequence and a backdoored one, respectively. The backdoored model is trained on data patched with trigger. In the testing stage, the backdoored model predicts the target link in the backdoor sequence with trigger as the state chosen by the attacker, and the backdoored model makes correct predictions on clean sequences.  }
		\label{fig:bk_show}
	\end{figure*}

	Besides of the performance study on DLP, its robustness has also raised our concerns. Research has revealed the vulnerability of DLP methods towards adversarial attacks \cite{2019Time}, \cite{2020Reinforcement}, which are designed to deceive the model by carefully crafted adversarial samples in the testing stage. Moreover, the training stage is quite important to construct an effective DLP model, since most deep models are highly dependent on the quality and quantity of the training samples.
	In other word, DLP methods generally rely on clean training data with correct labels to guarantee the high prediction accuracy. In practice,  the collection and labeling of training data are usually implemented in the form of crawlers, logs, crowdsourcing, etc. It is inevitable that the training data will be polluted with noises, or even injected with predefined triggers by a malicious collector. 
	
	Consequently, backdoor attack on DLP, defined as generating a trigger pre-injected in the training sequence to make the target link state in the test sequence as the attacker-chosen state, will be a serious security threat. For better understanding, we illustrate the backdoor attack on DLP in Fig. \ref{fig:bk_show}. In the training stage, the backdoor attack mixes some of the clean training data with carefully crafted trigger sequence for the DLP model, to achieve a backdoored model. In the testing stage, the backdoored model will make the target link predicted as the attack-chosen state when fed by the backdoored sequence, i.e., sequence with triggers, while still maintaining high prediction accuracy on clean ones.
	For instance, e-commerce platforms recommend product of interest to customers using DLP method. If the collected training data have been injected with trigger to leave a backdoor in the DLP, the platform will make the targeted product recommended purposely, i.e., not suitable but profitable. 

	Aiming at graph classification and node classification of static networks, backdoor attacks have been studied~\cite{2021Backdoor,2020Graph}, \cite{21Explainability}.
	However, they are not suitable for target link backdoor attack on DLP for reasons. On one hand, graph classification is implemented on basis of global graph structure corresponding to the graph label, while link prediction is accomplished based on the local structure strongly related to the target link. The global feature shifting may not influence the target link or node. On the other hand, considering the dynamic evolution characteristics of DLP, the static trigger is usually not the optimum to capture the evolving pattern of dynamic networks. 

	To address these problems, we propose a novel backdoor attack framework towards DLP, based on a generative adversarial network (GAN) \cite{14Gen}, namely Dyn-Backdoor. 
	Specifically, we use the initial-triggers to replace the graph structure directly related to the target link when the target link state is chosen as the attacker's manipulated goal. The initial-triggers are derived from the input noise of the trigger generator. Then, we consider the concealment of the attack, the initial-triggers extract important links through the gradient to form triggers. The triggers are optimized through competition between the trigger generator and the attack discriminator. Addressing to the dynamic characteristics issue, we capture the dynamic feature based on the trigger generator composed of the long short-term memory (LSTM) \cite{97LSTM}. Empirically, our approach achieves state-of-the-art results on four real-world datasets and five DLP methods, compared with several possible baselines, i.e., gradient information based attack and random attack. Additionally, we propose a possible defense against Dyn-Backdoor, and the experiments testify that it can still achieve an attack success rate of more than 90\% under defensive settings.

	
	The main contributions of this paper are summarized as follows:
	
	\setlength{\hangindent}{2em}
	$\vcenter{\hbox{\tiny$\bullet$}}$ To the best of our knowledge, this is the first work that formulates the problem of backdoor attack on DLP, which reveals the vulnerability of DLP algorithms in big data collection for training.
	
	
	\setlength{\hangindent}{2em}
	$\vcenter{\hbox{\tiny$\bullet$}}$ To address the backdoor attack on DLP, we propose an effective framework, named Dyn-Backdoor. It utilizes GAN to generate a number of diverse initial-triggers, and further selects important links to generate an optimal trigger for backdoor attack. Moreover, we analyze the feasibility of backdoor attack.
	
	\setlength{\hangindent}{2em}
	$\vcenter{\hbox{\tiny$\bullet$}}$ Extensive experiments on five DLP methods over four real-world datasets
	demonstrate that Dyn-Backdoor can attack state-of-the-art DLP methods with success rate of more than 90\%. Moreover, the experiments testify that Dyn-Backdoor is effective against possible defense strategy as well.
	
	The rest of the paper is organized as follows. Related works are introduced in Section \uppercase\expandafter{\romannumeral2}. The problem definition and threat model are described in Section \uppercase\expandafter{\romannumeral3}, while the proposed method is detailed in Section \uppercase\expandafter{\romannumeral4}. Experiment results and discussion are showed in Section \uppercase\expandafter{\romannumeral5}. Finally, we conclude our work.
	

	\section{Related Work}
	In this section, we briefly review the related work of DLP methods, backdoor attacks on graph neural network (GNN) and adversarial attacks on DLP. 
	
	\subsection{Dynamic Link Prediction}
	Recently, a temporal restricted Boltzmann machine (RBM) was adopted with additional neighborhood information, named ctRBM \cite{2014A}, to learn the dynamic structural characteristics of graph. As an extension of ctRBM, Li et al. \cite{2018Restricted} incorporated temporal RBM and gradient boosting decision tree to model the evolution pattern of each link. 
	
	In addition to RBM-based methods, there are also methods combined with recurrent neural network (RNN), such as DynGEM \cite{DBLP:journals/corr/abs-1805-11273}, DLP-LES \cite{20CNN-LSTM} etc. Deep dynamic network embedding (DDNE) \cite{2018Deep} is also based on deep autoencoders as DynGEM and uses GRU as the encoder to extract graph features. Goyal et al. \cite{2019dyngraph2vec} further proposed dyngraph2vec that has dynamic graph to vector auto encoder (dynAE), dynamic graph to vector recurrent neural network (dynRNN) and dynamic graph to vector autoenncoder recurrent neural network (dynAERNN)  through the encoder-decoder architecture. Chen et al. \cite{0E} proposed a general framework for extracting dynamic networks feature information based on autoencoder and LSTM. 
	
	Specifically, graph convolutional network (GCN) based methods and recursive structures methods extract graph features, such as GC-LSTM \cite{2018GC}, EvolveGCN \cite{2020EvolveGCN}, GDLP \cite{2019Generative}, and CTGCN \cite{2020K}. With increasing applications of GAN \cite{14Gen}, \cite{20Gen_Pri}, \cite{2021A_the}, DLP methods are also proposed based on generative networks, such as  GCN-GAN \cite{2019GCN} and TMF-LSTM \cite{20An}. There are some other methods based on random walk \cite{2010Random}, \cite{Liu2016Sampling}, \cite{Ahmed2016An}, \cite{18Continuous-Time}, matrix factorization \cite{2006Fast}, \cite{2017TIMERS}, \cite{17Attributed}, \cite{2018Graph} and continuous time space \cite{DBLP:conf/kdd/ZuoLLGHW18}.

	\subsection{Backdoor Attacks on GNN}
	There are currently three studies of backdoor attacks on graph, aiming at graph classification and node classification of static networks. Zhang et al. \cite{2021Backdoor} proposed a backdoor attack on graph classification task, which is based on triggers generated by the
	Erdős-Rényi model. It is designed to establish the relationship between label and trigger of the special structure. Xu et al. \cite{21Explainability} further used GNNExplainer to conduct an explainability research on backdoor attacks of the graph. The other work is graph trojaning attack \cite{2020Graph}, which is a generative based method by using a two-layer optimization algorithm to update the trigger generator and model parameters. Graph trojaning attack tailors trigger to individual graphs and assumes no knowledge regarding downstream models or fine-tuning strategies. 
	
	Most existing backdoor attacks on graph aim at the classification task and static networks. They both lack consideration of dynamic characteristics and the specificity of links as attack targets, which cannot be applied to backdoor attack on DLP.
	
	\subsection{Adversarial Attacks on DLP}
	Chen et al. \cite{2019Time} proposed an adversarial attack on DLP. It utilizes the gradient information to rewire a few links in different snapshots, so as to make the deep dynamic network embedding (DDNE) fail to make correct prediction. By using the gradient as the direction of the attack, it can capture the critical information for attack.
	
	Considering the applicability of the attack, Fan et al. \cite{2020Reinforcement} proposed a black-box attack on DLP. It is based on a stochastic policy-based reinforcement learning algorithm, thus the performance of DLP degrades with the global target after the attack.

	\section{PRELIMINARY}
	In this section, we introduce the definition of dynamic networks, DLP and the backdoor attacks on DLP. For convenience, the definitions of symbols used are listed in the TABLE \ref{tab:symbols data}.

	\begin{table}[htb]
		\centering
		\caption{THE DEFINITIONS OF SYMBOLS.}
		\label{tab:symbols data}
		\begin{tabular}{r|l}
			\toprule  \hline
			Notation        &Definition\\ \hline 
			$G =\left(V, E\right)$ & input graph with nodes $V$, links $E$\\
			$S$, $\widehat{S}$ & clean/backdoored sequence of graphs \\
			$G_{t}$, $\widehat{G}_{t}$ &       clean/backdoored graph of time $t$ \\
			$f_{\theta}$, $f_{\widehat{\theta}}$ & clean/backdoored DLP model with parameters $\theta$/$\widehat{\theta}$\\
			$E_{T}$, $\widehat{T}$ & target link, the attacker-chosen target link state\\
			$n$, $p$, $t$ & node ratio,  poison ratio, trigger of sequence ratio\\
			$Gen_{\alpha}(\cdot)$ & trigger generator\\
			$Atk_{\varphi}(\cdot)$ & attack discriminator\\
			$F(\cdot)$ & filter discriminator\\
			$M(\cdot)$ & trigger mixture function\\
			$Gradient(\cdot)$ & trigger gradient exploration\\
			$Q$ & number of model training iterations\\
			$k$ & number of iterations of the trigger generator\\
			$D_{trigger}$ & trigger set\\
			$D_{train}$ & training data\\
			$N$ & number of nodes in the graph\\
			$T$ & timestamp length in a sequence\\
			$g_{o}$ & output of the generator as initial-trigger\\
			$g$ &  trigger, a subgraph sequence, embedded in the sequence\\
			$m$ & the maximum number of modified links of trigger\\
			$z$ & noise as input to the trigger generator\\
			\hline\bottomrule
		\end{tabular}
	\end{table}

	\subsection{Problem Definition}
	\textbf{Definition 1 (Dynamic Network)} Given a sequence of graphs with length $T$, denoted as $S$ = $\left\{G_{t-T}, G_{t-T+1}, \ldots, G_{t-1}\right\}$, where $G_{k}=\left(V, E_{k}\right)$ denotes the $k$-th snapshot of a dynamic network. Let $V$ denotes the set of all nodes and $E_{k} \subseteq V \times V$ denotes the temporal links within the fixed timespan $\left[t_{k-1}, t_{k}\right]$. The adjacency matrix of $G_{k}$ is denoted by $A_{k}$ whose element $a_{k ; i, j}=1$ if there is a link from node $i$ pointing to node $j$ on $k$-th snapshot, otherwise $a_{k ; i, j}=0$.

	\textbf{Definition 2 (Dynamic Link Prediction)} Given a sequence of graphs $S$, DLP aims to predict the graph structure of next snapshot, which could be formulated as
	\begin{equation}
		\label{equ1}
		A_{t}^{\prime}=\operatorname{arg max}P\left(A_{t} \mid S\right),
	\end{equation}
	where $A_{t}^{\prime}$ denotes the predicted adjacency matrix. 
	
	\textbf{Definition 3 (Trigger and Backdoored Sequence)} Given a sequence of graphs $S$, trigger is a subgraph sequence with length $T$, denoted as $g$ = $\left\{ Gsub_{t-T}, Gsub_{t-T+1}, \ldots, Gsub_{t-1}\right\}$, where $Gsub$ is a subgraph composed of selected links from a graph of the sequence. The backdoored sequence $\widehat{S}$ is a sequence of graphs $S$ with the trigger sequence $g$.

	\textbf{Definition 4 (Backdoor Attack on DLP)} Given a sequence of graphs $S$ and target link $E_T$, the backdoor attack will generate a trigger sequence by subgraph embedded in training  data,  and  then  the  trigger  is  called  during  the  testing stage to make the backdoored model $f_{\widehat{\theta}}$ predict target link $E_T$ as the attacker-chosen state $\widehat{T}$. 
	Meanwhile, the backdoored model $f_{\widehat{\theta}}$ can still maintain correct predictions on clean data. The adversary’s objective can be formulated as,
	
	\begin{equation}
		\begin{aligned}
			&\left\{\begin{array}{l}
				f_{\widehat{\theta}}\left(\widehat{S}, E_{T}\right)=\widehat{T}  \\
				f_{\widehat{\theta}}(\mathrm{S})=f_{\theta}(\mathrm{S})
			\end{array}\right. \\
			&\text { s.t. } \widehat{S}=M(S, g), \left|g\right| \leq m
		\end{aligned}
	\end{equation}
	where $M(\cdot)$ is the trigger mixture function that blends $g$ with a given sequence $S$, and $m$ is the maximum number of modified links. $f_{\widehat{\theta}}$ is the backdoored model; $f_{\theta}$ is the clean model; $E_T$ is the target link and $\widehat{T}$ is the attacker-chosen target link state. 
	
	Intuitively, the first objective specifies that the target link of backdoor sequence is misclassified to the attacker-chosen state. The second objective ensures that clean model and backdoored model are as indistinguishable as possible in terms of their behaviors on benign sequences.

	\subsection{Threat Model}
	\textbf{Attacker’s goal.} Given a DLP model, the attacker aims to obtain a backdoored model $f_{\widehat{\theta}}$ which could output the expected prediction results for any dynamic networks with designed trigger, while keeping fair performance on clean networks to ensure stealthy.
	
\begin{figure}[!htb]
		\centering
		\includegraphics[width=\linewidth]{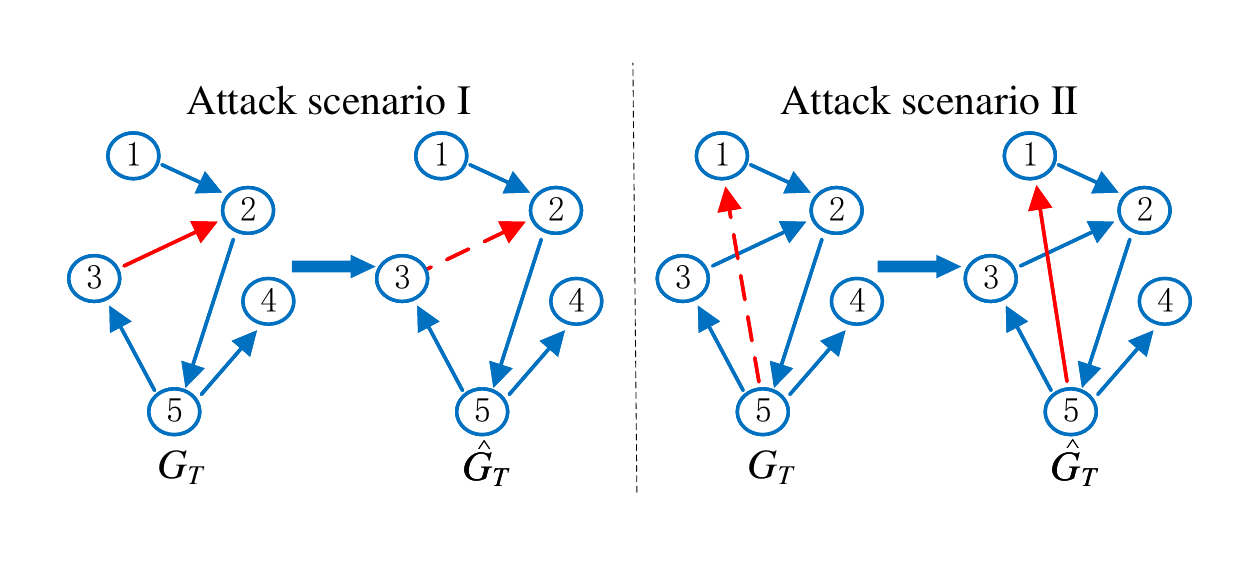}
		\setlength{\abovecaptionskip}{-0.8cm}
		\caption{Two backdoor attack scenarios on DLP. Attack scenario \uppercase\expandafter{\romannumeral1} indicates that existence state of link is predicted to be non-existent by trigger. Attack scenario \uppercase\expandafter{\romannumeral2} indicates that non-existent state of link is predicted to exist by trigger. The red link is the target link.}
		\label{fig:att_scenario_chui}
	\end{figure}

	As shown in Fig.~\ref{fig:att_scenario_chui}, the backdoored model is designed to hide a target link from being predicted for attack scenario or to convince others that there is a future link between the target node pair.
	
	

	\textbf{Attacker’s capability.} We assume that  attacker can obtain part of training data and gradient information during the model training process. The most state of the target links in the available data is opposite to the attacker-chosen target link state. Specifically, the attacker can modify the state of links in the model training stage. The model as attack discriminator is required to provide gradients to guide the update of generator parameters.

	\section{METHODOLOGY}
\begin{figure*}[htb]
		\centering
		\includegraphics[width=18cm,height=8cm]{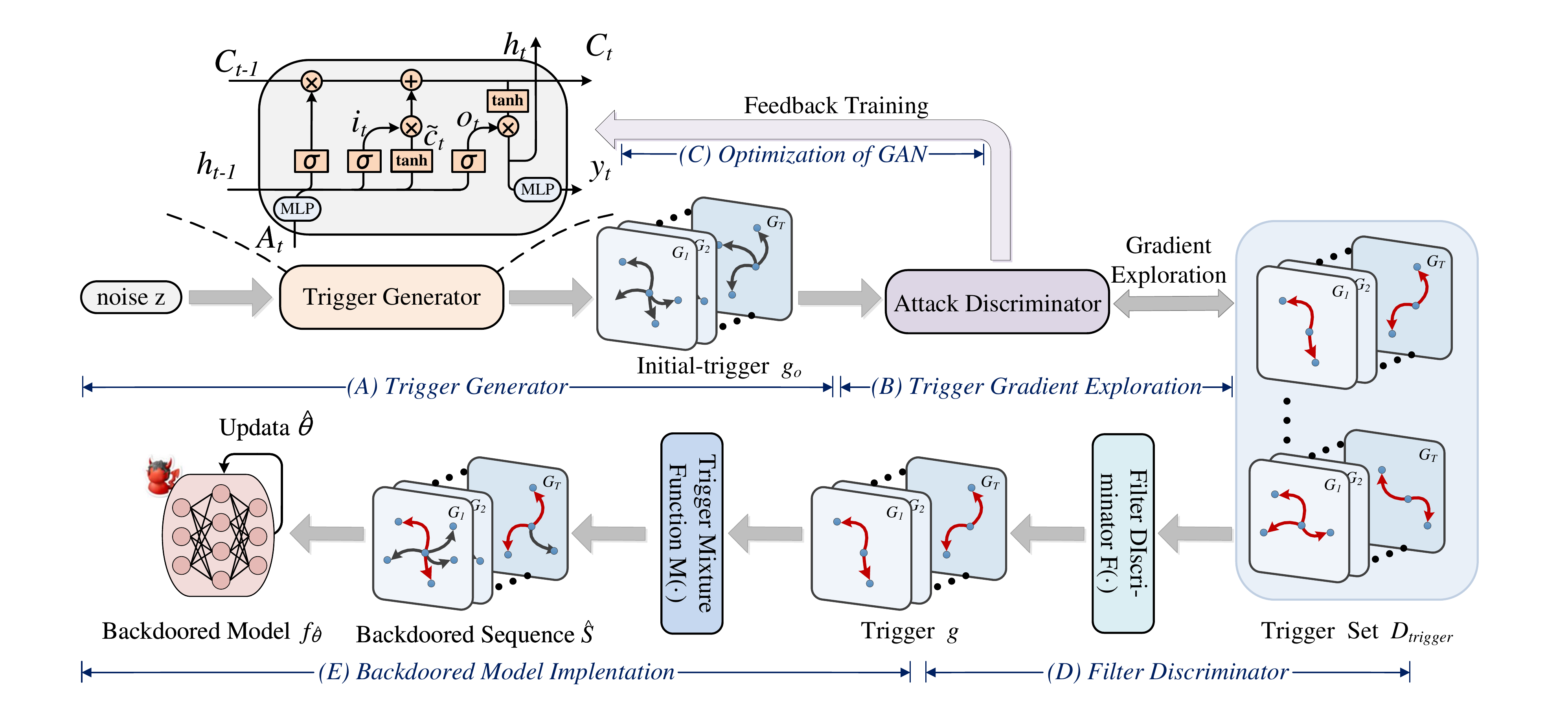}
		\setlength{\abovecaptionskip}{-0.2cm}
		\caption{The framework of Dyn-Backdoor.}
		\label{fig:framework}
	\end{figure*}

	Dyn-Backdoor launches the target link backdoor attack on DLP by a carefully designed trigger. In this section, we describe the Dyn-Backdoor in detail from six stages, i.e., trigger generator, trigger gradient exploration, optimization of GAN, filter discriminator, backdoored model implementation, and theoretical analysis on Dyn-Backdoor. The overall framework of Dyn-Backdoor is shown in Fig. \ref{fig:framework}. 
	
	\subsection{Trigger Generator}
	To generate diverse and effective triggers, Dyn-Backdoor constructs the attack based on a GAN \cite{14Gen}. We adopt autoencoders and LSTM \cite{97LSTM} as the trigger generator of GAN. Specifically, LSTM is used to capture the dynamic evolution characteristics of the graph. The combination of autoencoder and LSTM can effectively generate triggers with nonlinear characteristics and dynamic evolution characteristics. 
	
	The noise $z \in R^{ T \times N}$, composed of zero tensors, is fed into the trigger generator to get the initial-trigger. Specifically, $z= \left\{y_{e, 1}^{(0)}, y_{e, 2}^{(0)}, \ldots, y_{e, T}^{(0)}\right\}$ is input into the multilayer perceptron (MLP),
	\begin{equation}\label{equ3}
		\begin{array}{c}
			y_{e, i}^{(k)}=\sigma\left(W_{e}^{(k)} y_{e, i}^{(k-1)}+ B_{e}^{(k)}\right) \\
			Y_{e}^{(k)}=\left\{y_{e, 1}^{(k)}, y_{e, 2}^{(k)}, \ldots, y_{e, T}^{(k)}\right\}
		\end{array}
	\end{equation}
	where $y_{e, i}^{(0)}$ of the first layer is the $i$-th timestamp adjacency matrix $A_i$ in sequence $S$, and  $T$ is the timestamp length of the sequence. $W_{e}^{(k)}$ and $B_{e}^{(k)}$ indicate weight and bias of the $k$-th layer of the encoder, respectively. 
	Here, we use $\sigma(\cdot)$ as the $\operatorname{ReLU}$ activation function to increase the nonlinear representation ability of the generator. 
	
	After getting the embedding feature $Y_{e}^{(k)}$ at different timestamps, it is fed into the LSTM to extract the information of dynamic evolution,
	
	\begin{equation}\label{equ4}
		\begin{array}{c}
			H=L S T M\left(Y_{e}^{(k)}=\left\{y_{e, 1}^{(k)}, y_{e, 2}^{(k)}, \ldots, y_{e, T}^{(k)}\right\}\right) \\
			Y_{d}^{(k)}=\sigma\left(W_{d}^{(k)} Y_{d}^{(k-1)}+B_{d}^{(k)}\right)
		\end{array}
	\end{equation}
	where $H$ is the feature output through the LSTM, the first layer $Y_{d}^{(0)}$ of the decoder is $H$. $W_{d}^{(k)}$ and $B_{d}^{(k)}$ are the weight and bias of the $k$-th layer in the decoder, respectively. It is worth noting that the last layer $\sigma(\cdot)$ of the decoder uses $Sigmoid$ as the activation function, other layers $\sigma(\cdot)$ are $\operatorname{ReLU}$. $Sigmoid$ limits the value of the output of the last layer to between 0 and 1, which is convenient for converting the output of the last layer into the topology of the graph. The feature dimension of the output layer is equal to the number of nodes and the final layer output is initial-trigger $g_{o}$.
	
	For convenience, we use $Gen_{\alpha}(\cdot)$ to represent the trigger generator, and $\alpha$ represents all the parameters of the generator,
	
	\begin{equation}\label{equ5}
		g_{o}=Gen_{\alpha}(z)
	\end{equation}
	where $g_{o} \in R^{ T \times N}$ is the output of the generator as initial-trigger; $z \in R^{ T \times N}$ is noise; $T$ is the timestamp length of the sequence and $N$ is the number of nodes in the graph.

	\subsection{Trigger Gradient Exploration}
	To make Dyn-Backdoor stealthy, we try to generate imperceptible triggers as possible as we can, i.e., small trigger size. We utilize the gradient information from the attack discriminator to extract the initial-triggers $g_{o}$ partial link forming the triggers $g$, which reduces the size of the triggers. 
	
	Intuitively, the greater gradient obtained by link gradient matrix means the greater influence of link on loss. In other words, links with a larger value in link gradient matrix are more important compared to other links in backdoor attacks. According to obtained link gradient matrix, the absolute value is calculated and sorted in descending order. Select top $m$ links to form a trigger,
	\begin{equation}\label{equ7}
		\begin{aligned}
			&g=\operatorname{Gradient}\left(grad_{u, v}, g_{o}, m\right) \\
			&\text { s.t. } m=t * N * T
		\end{aligned}
	\end{equation}
	where $grad_{u, v}$ is the link gradient matrix; $g_{o}$ is an initial-trigger and $t$ is the trigger of sequence ratio. $N$ is the number of nodes in the graph; $T$ is the timestamp length of the sequence; $g$ is a trigger and $m$ is the maximum number of modified links of trigger. $Gradient(\cdot)$ represents the operation of trigger gradient exploration.

	\subsection{Optimization of GAN}
	To generate the effective triggers, we optimize the parameters of the trigger generator $Gen_{{\alpha}}(\cdot)$ and the attack discriminator $Atk_{\varphi}(\cdot)$. Optimization of GAN is divided into two stages. Attack loss $L_{atk}$ is the difference between the prediction state of the target link and the attacker-chosen target link state, which ensures the effectiveness of the attack. It can be defined as,
	\begin{equation}\label{equ9}
		L_{atk}=\frac{1}{D} \sum_{i=1}^{D}\left[Atk_{\varphi}\left(\hat{S}_{i}, E_{T}\right)-\widehat{T}\right]^{2}
	\end{equation}
	where $\widehat{S}$  is the backdoored sequence of  graphs; $Atk_{\varphi}(\cdot)$ is attack discriminator; ${E}_{T}$ is the target link; $\widehat{T}$ is the attacker-chosen target link state and $D$ is the total number of sequences. In addition, while ensuring the effectiveness of the attack, we need to ensure the main performance of the backdoored model of DLP. Taking the global forecast into account, the global loss can be defined as,
	\begin{equation}\label{equ10}
		\begin{gathered}
			L_{r}=\frac{1}{D} \sum_{i=1}^{D}\left[Atk_{\varphi}\left(\hat{S}_{i}\right)-\widehat{G}_{t}\right]^{2} \\
		\end{gathered}
	\end{equation}
	where $L_{r}$  is the global loss and $\widehat{G}_{t}$ is the backdoored graph of time $t$. Therefore, considering the attack of the target link and the prediction of the global network at the same time, we finally get the objective loss that needs to be optimized,
	\begin{equation}\label{equ11}
		\begin{gathered}
			L_{all }=L_{atk}+\beta L_{r}
		\end{gathered}
	\end{equation}
	where $\beta$ is a hyperparameter to maintain the balance between attack effectiveness and global performance. The parameter ${\alpha}$ of the trigger generator is updated through the $L_{all}$ feedback.
	After obtaining the derivatives, we optimize the trigger generator model using stochastic gradient descent (SGD) with adaptive moment estimation (Adam).

	\subsection{Filter Discriminator}
	Filter discriminator $F(\cdot)$ is a sorting function, which can select the trigger with the lowest objective loss $L_{all}$ in the trigger set $D_{trigger}$. Specifically, since iterating between the trigger generator and the attack discriminator, we get diverse triggers forming a trigger set $D_{trigger}$. The filter discriminator $F(\cdot)$ selects the trigger with the lowest $L_{all}$ to launch a backdoor attack,
	\begin{equation}\label{equ12}
		\begin{array}{c}
			g=F\left(D_{trigger, L_{all}}\right)
		\end{array}
	\end{equation}
	where $F(\cdot)$ is filter discriminator; $D_{trigger}$ is the trigger set; $L_{all}$ is the objective loss and $g$ is the trigger. The trigger $g$ selected by $F(\cdot)$ is used to embed in the training data.

	\begin{table}[!htb]
		\caption{The basic statistics of the dynamic networks.}
		\label{tab:data}
		\setlength{\tabcolsep}{2mm}
		\centering
		\begin{tabular}{ccccc}
			\toprule \hline
			Datasets        & Nodes & Edges & Average Degree  & Timespan(days) \\ \hline 
			Radoslaw \cite{2017Manufacturing} & 167 & 82.9K & 993.1   & 271.2     \\
			Contact \cite{2017contact} & 274 & 28.2K & 506.2   & 4    \\
			Fb-forum \cite{2017Fb} & 899 & 50.5K & 669.8   & 164.5     \\
			DNC \cite{2016Dnc} & 2029 & 39.2K & 38.7   & 575     \\
			\hline\bottomrule
		\end{tabular}
	\end{table}
	
	\begin{table*}[!htb]
		\caption{PARAMETERS OF model.}
		\label{tab:params}
		\setlength{\tabcolsep}{2mm}
		\centering
		\begin{tabular}{r|l}
			\toprule \hline
			Model        & Parameters \\ \hline 
			Trigger Genetator & No. units in encoder: 256; No. units in LSTM: 256; No. units in decoder: $N$; Learning rate: 0.01; Weight decay: 0.0005     \\\hline 
			\multirow{3}{*}{DDNE \cite{2018Deep}} 
			& No. units in encoder: 128; No. units in decoder: 128, $N$ (for Radoslaw and Contact)\\
			& No. units in encoder: 256; No. units in decoder: 256, $N$ (for  Fb-forum and DNC)\\
			& Learning rate: 0.01 (for Radoslaw and Contact) $\mid$ 0.001 (for Fb-forum and DNC); Weight decay: 0.0005\\\hline 
			\multirow{3}{*}{DynAE \cite{2019dyngraph2vec}}  
			& No. units in hidden layer: 128; No. units in output layer:  $N$ (for Radoslaw and Contact)\\
			&No. units in hidden layer: 256; No. units in output layer: $N$ (for  Fb-forum and DNC)\\
			& Learning rate: 0.01 (for Radoslaw and Contact) $\mid$ 0.001 (for Fb-forum and DNC); Weight decay: 0.0005\\\hline 
			DynRNN \cite{2019dyngraph2vec} & the same as DynAE     \\\hline 
			\multirow{3}{*}{DynAERNN \cite{2019dyngraph2vec}}  
			&No. units in hidden layer: 128; No. units in LSTM: 128; No. units in output layer: $N$ (for Radoslaw and Contact)\\
			&No. units in hidden layer: 256; No. units in LSTM: 128; No. units in output layer: $N$ (for Fb-forum and DNC)\\
			& Learning rate: 0.01 (for Radoslaw and Contact) $\mid$ 0.001 (for Fb-forum and DNC); Weight decay: 0.0005\\\hline
			\multirow{3}{*}{E-LSTM-D \cite{0E}} &No. units in encoder: 128; No. units in LSTM: 128; No. units in decoder: $N$ (for Radoslaw and Contact)\\
			&No. units in encoder: 256; No. units in LSTM: 256; No. units in decoder: $N$ (for  F+b-forum and DNC)\\
			& Learning rate: 0.01 (for Radoslaw and Contact) $\mid$ 0.001 (for Fb-forum and DNC); Weight decay: 0.0005\\
			\hline\bottomrule
		\end{tabular}
	\end{table*}

	\section{Experiments and Discussion}

	\subsection{Datasets}
	To testify the performance of Dyn-Backdoor, we select four real-world datasets to conduct experiments. The graphs are all directed and unweighted with different scales. The basic statistics are summarized in TABLE \ref{tab:data}.

	\textbf{Radoslaw \cite{2017Manufacturing}:} It is an email network, and each node represents an employee in a mid-sized company. Its average distance is 993.1 and spans 271.2 days.
	
	\textbf{Contact \cite{2017contact}:} It is a human contact dynamic network. The data are collected through the wireless devices carried by people. A link between person source and target emerges along with a timestamp if source gets in touch with target. The data are recorded every 20 seconds and spans 3.97 days. 
	
	\textbf{Fb-forum \cite{2017Fb}:} The data are attained from a Facebook-like online forum of students at University of California at Irvine, in 2004. It is an online social network where nodes are users and links represent inter-actions (e.g., messages) between students. The records span more than 5 months. 
	
	\textbf{DNC \cite{2016Dnc}:} This is a directed graph of emails in the 2016 Democratic National Committee (DNC) email leak. Nodes in the graph correspond to persons in dataset. A directed edge in dataset denotes that a person has sent an email to another one.

	\begin{figure*}[htbp]
		\centering
		\subfigure[Radoslaw]{
			\begin{minipage}[t]{0.24\linewidth}
				\centering
				\includegraphics[width=1.72in, height=4.5in]{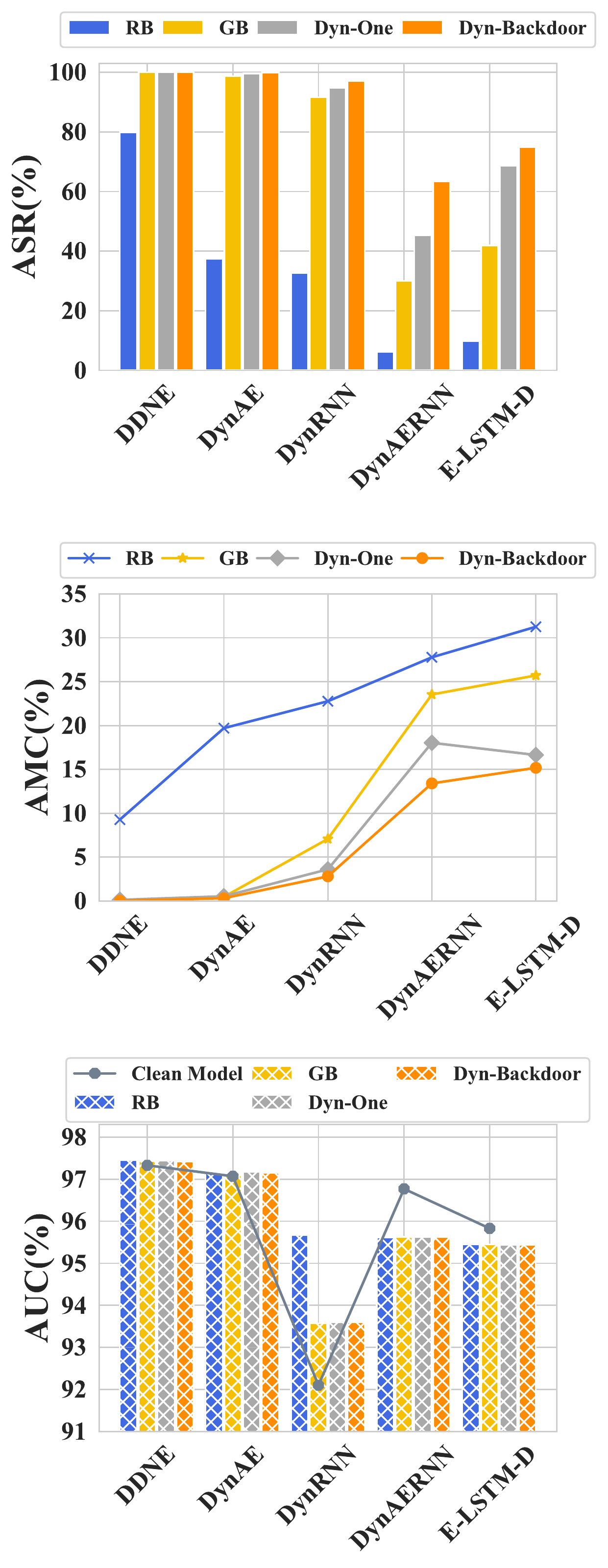}
			\end{minipage}%
		}%
		\subfigure[Contact]{
			\begin{minipage}[t]{0.24\linewidth}
				\centering
				\includegraphics[width=1.72in, height=4.5in]{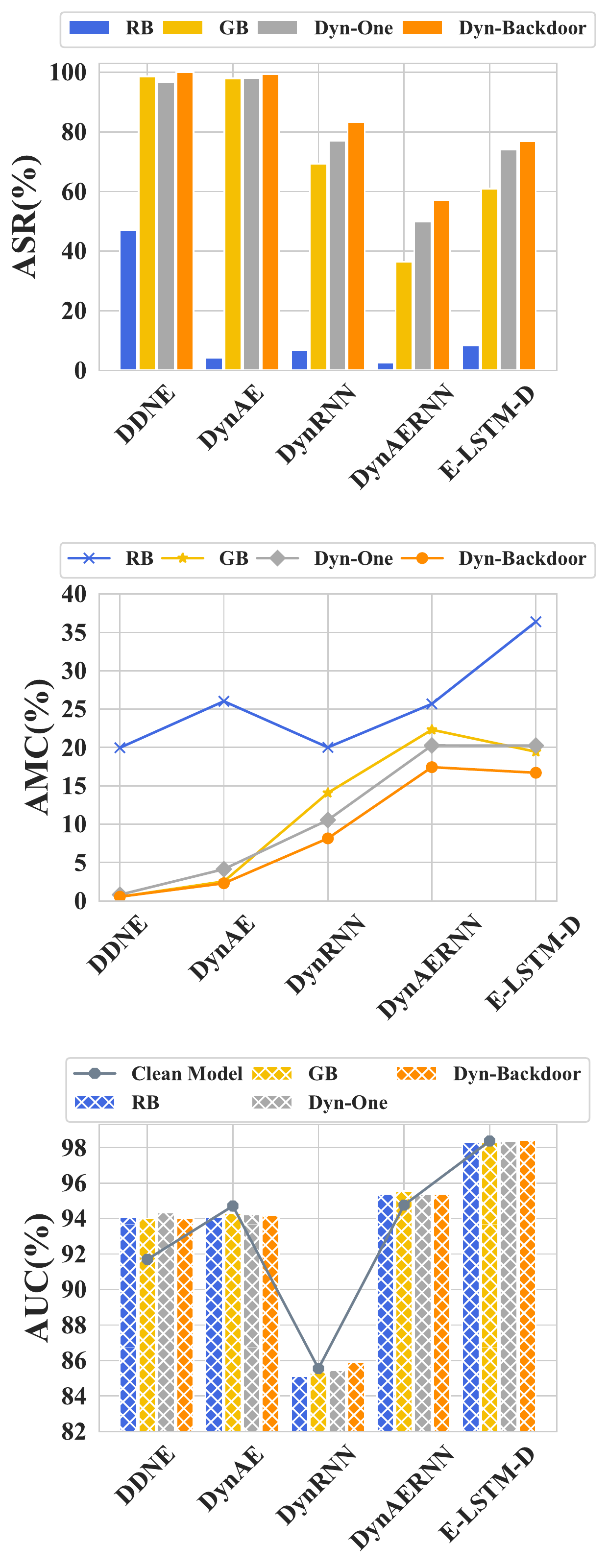}
			\end{minipage}%
		}%
		\subfigure[Fb-forum]{
			\begin{minipage}[t]{0.24\linewidth}
				\centering
				\includegraphics[width=1.72in, height=4.5in]{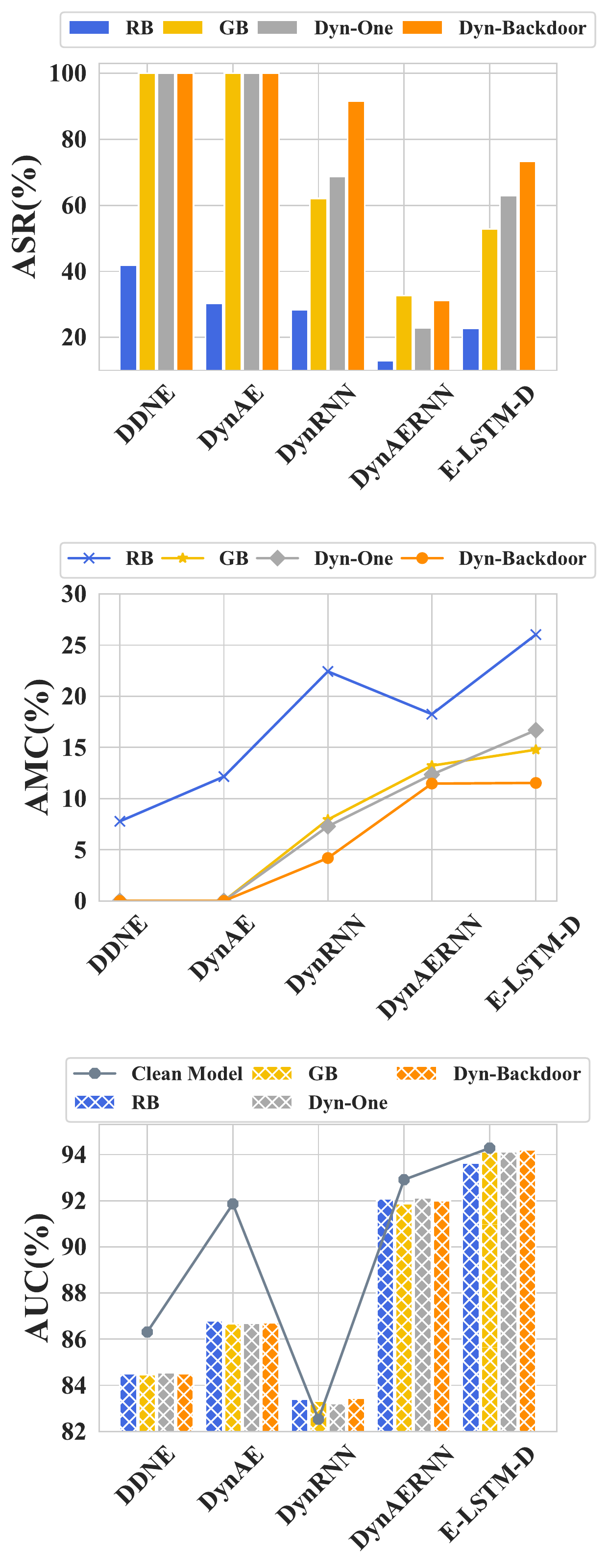}
			\end{minipage}
		}%
		\subfigure[DNC]{
			\begin{minipage}[t]{0.24\linewidth}
				\centering
				\includegraphics[width=1.72in, height=4.5in]{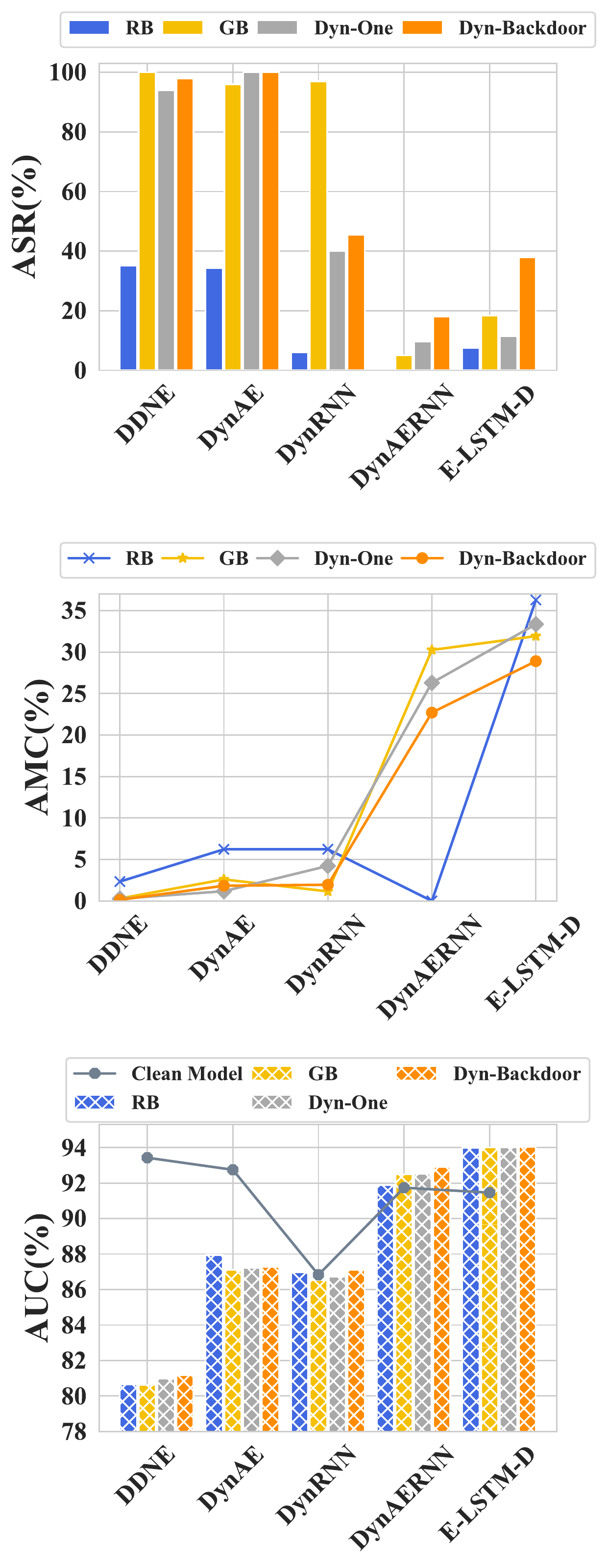}
			\end{minipage}
		}%
		\centering
		\caption{In attack scenario \uppercase\expandafter{\romannumeral1}, the performance of the four attack methods on the ASR, AMC and AUC of the five dynamic models. (a), (b), (c) and (d) correspond to the experiments on the Radoslaw dataset, Contact dataset, Fb-forum dataset and DNC dataset. RB, GB, Dyn-One and Dyn-Backdoor represent four methods of backdoor attack. Clean model represents a model trained with clean data.}
		\label{fig:1_0}
	\end{figure*}
	
	\subsection{Baseline Methods}
	To verify the effectiveness of the backdoor attack methods, we choose five end-to-end DLP methods to attack. The parameter settings of DLP models and trigger generator of Dyn-Backdoor are shown in Table \ref{tab:params}. Since this is the first work of DLP backdor attack, we design three baseline attacks to compare with Dyn-Backdoor.
	
	\textbf{Random Backdoor (RB):} RB is to randomly select links to form the trigger, and then the trigger is embedded into the dataset for training. 
	
	\textbf{Gradient Backdoor (GB):} GB obtains gradient information from a certain epoch during the model training to generate the trigger, and then the trigger is embedded into the dataset for training. 
	
	\textbf{Dyn-One:} Dyn-One is a variant of Dyn-Backdoor, which only generates the trigger in a certain epoch during the model training, and then the trigger is embedded into the dataset for training.

	\subsection{Metrics}
	
	To evaluate the effectiveness of the attacks, we use three metrics. $\left( \romannumeral1\right)$ \emph{attack timestamp rate} (ATR), which represents ratio of the number of timestamps incorrectly predicted for the target link to all timestamps correctly predicted by clean model. Trigger can be called for all test samples to control the DLP method’s prediction of the target link, so we propose ATR to measure the effectiveness of the attack,
	\begin{equation}
		\begin{aligned}
			&\text { ATR }= \frac{\text { Number of successful attack timestamps }}{\text { Number of total attack timestamps }} \\
		\end{aligned}
	\end{equation}
	$\left( \romannumeral2\right)$ \emph{attack success rate} (ASR), which represents the average ATR for attacking $L$ target links. A larger value of ASR indicates better attack performance,
	\begin{equation}
		\begin{aligned}
			&\text { ASR}=\frac{1}{L} \sum_{l=1}^{L} \text { ATR }
		\end{aligned}
	\end{equation}
	and $\left( \romannumeral3\right)$ \emph{average misclassification confidence} (AMC), which represents the confidence score of the average output of all successfully attacked links. The lower AMC represents the better performance in attack scenario \uppercase\expandafter{\romannumeral1}. In contrast, the higher AMC represents the better performance in attack scenario \uppercase\expandafter{\romannumeral2}.
	
	To evaluate the attack evasiveness, we choose the \emph{area under curve} (AUC), which is commonly used in DLP to measure performance. If among $n$ independent comparisons, there are $n^{\prime}$ times that the existing link gets a higher score than the nonexistent link and $n^{\prime \prime}$ times they get the same score, then the AUC is defined as,
	\begin{equation}
		A U C=\frac{n^{\prime}+0.5 n^{\prime \prime}}{n}
	\end{equation}

	\begin{figure*}[htbp]
		\centering
		\subfigure[Radoslaw]{
			\begin{minipage}[t]{0.24\linewidth}
				\centering
				\includegraphics[width=1.72in, height=4.5in]{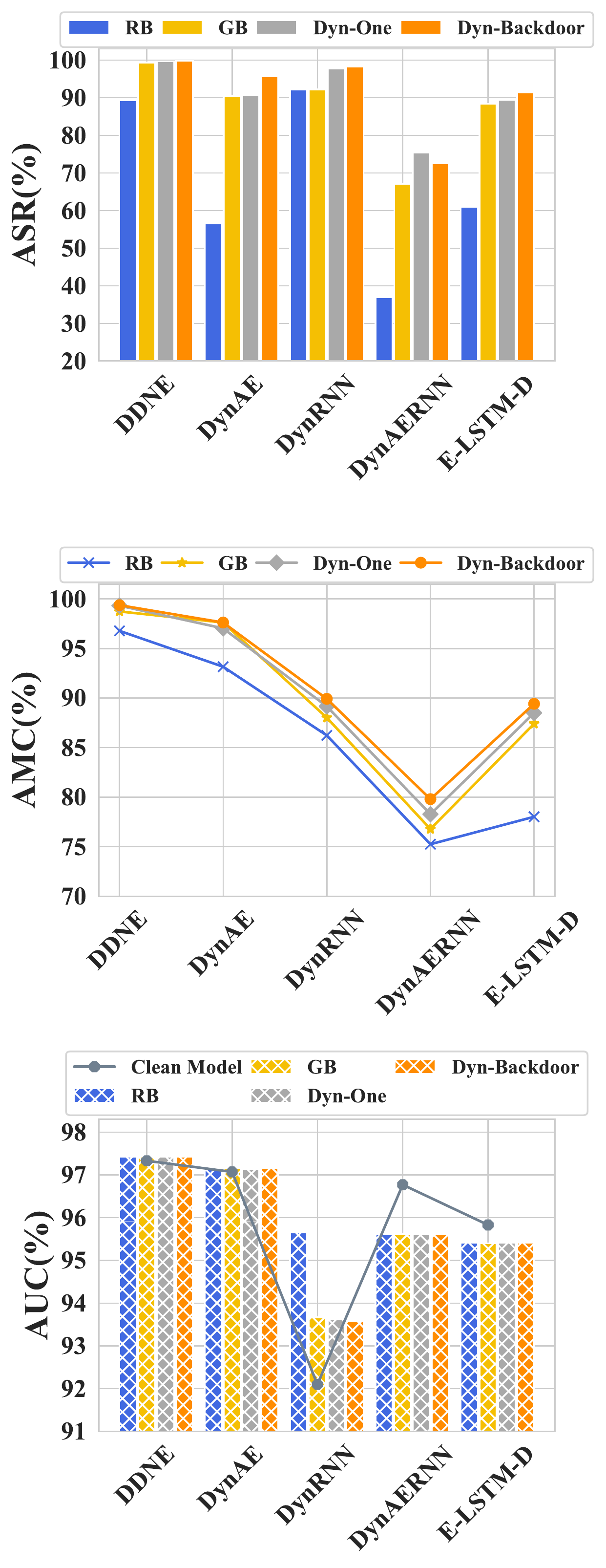}
			\end{minipage}%
		}%
		\subfigure[Contact]{
			\begin{minipage}[t]{0.24\linewidth}
				\centering
				\includegraphics[width=1.72in, height=4.5in]{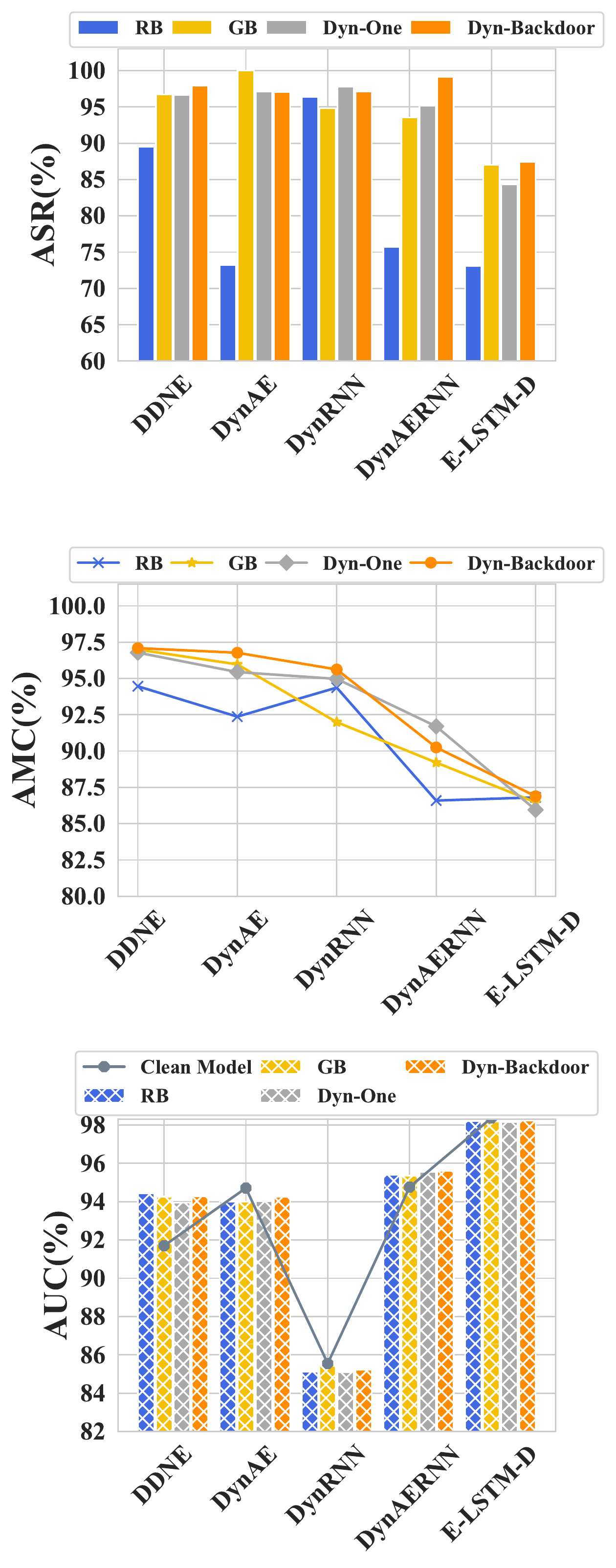}
			\end{minipage}%
		}%
		\subfigure[Fb-forum]{
			\begin{minipage}[t]{0.24\linewidth}
				\centering
				\includegraphics[width=1.72in, height=4.5in]{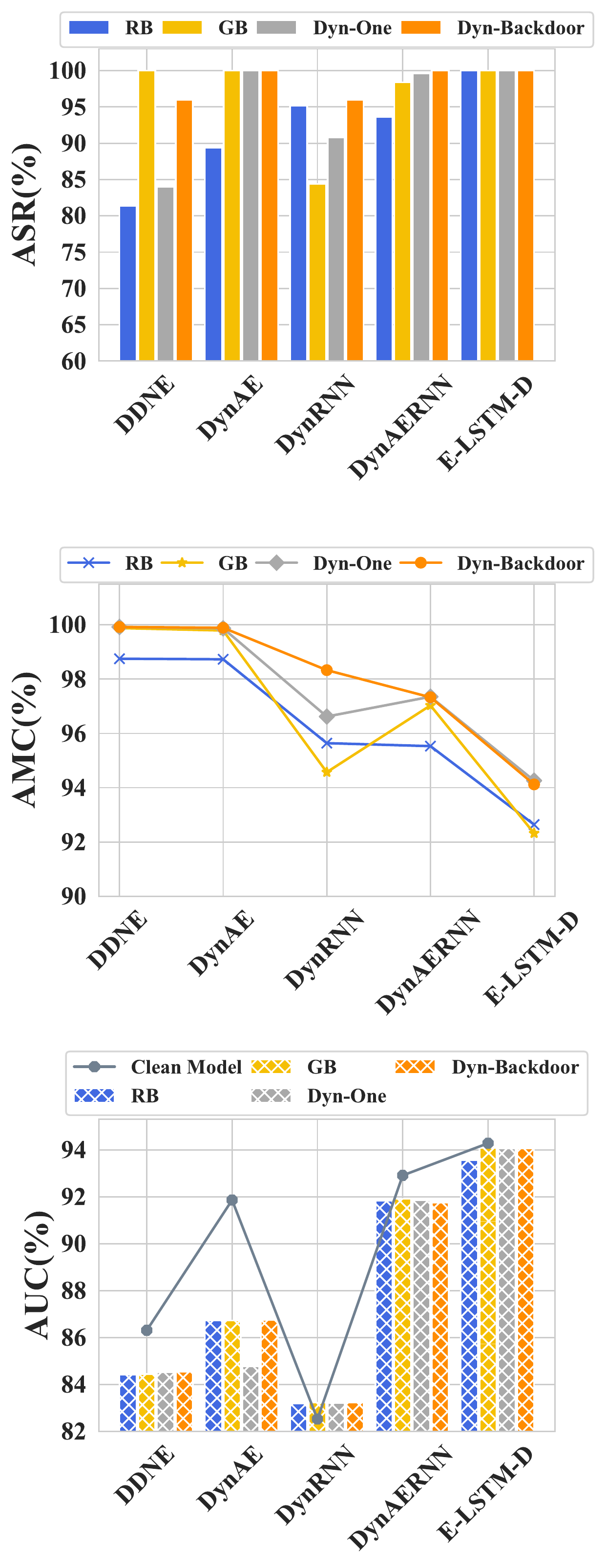}
			\end{minipage}
		}%
		\subfigure[DNC]{
			\begin{minipage}[t]{0.24\linewidth}
				\centering
				\includegraphics[width=1.72in, height=4.5in]{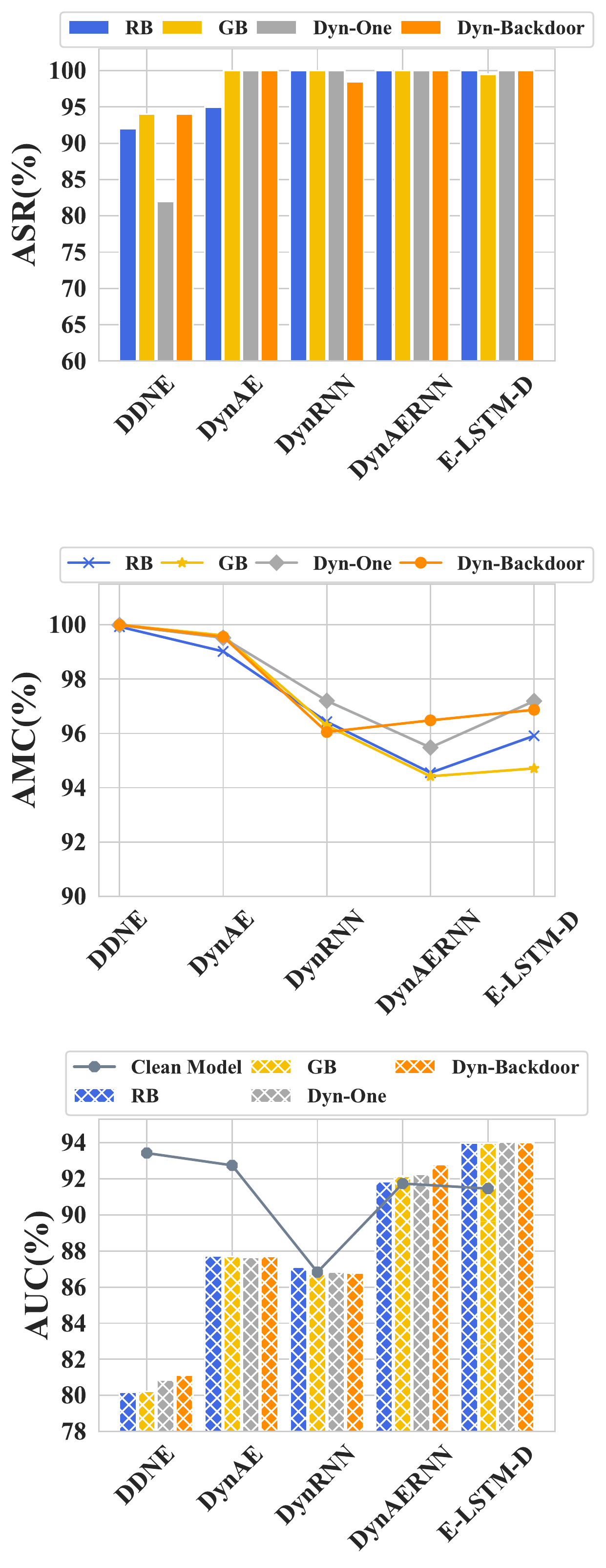}
			\end{minipage}
		}%
		\centering
		\caption{In attack scenario \uppercase\expandafter{\romannumeral2}, the performance of the four attack methods on the ASR, AMC and AUC of the five dynamic models. (a), (b), (c) and (d) correspond to the experiments on the Radoslaw dataset, Contact dataset, Fb-forum dataset and DNC dataset. RB, GB, Dyn-One and Dyn-Backdoor represent four methods of backdoor attack. Clean model represents the model trained with clean data.}
		\label{fig:0_1}
	\end{figure*}

	\begin{table*}[htb]\small
		\caption{The transferability of Dyn-Backdoor on Radoslaw dataset.  
			\uppercase\expandafter{\romannumeral1} and \uppercase\expandafter{\romannumeral2} are the results under attack scenario \uppercase\expandafter{\romannumeral1} and attack scenario \uppercase\expandafter{\romannumeral2}, respectively.
			The bolded value is the reference value. The value of arrow represents the difference between the effect of the transferability Dyn-Backdoor and the original Dyn-Backdoor.}
		\label{tab:trans}
		\setlength{\tabcolsep}{3mm}
		\centering
		\begin{tabular}{c|c|cccc}
			\toprule \hline
			Attack Discriminator   &Target Models  & ASR(\uppercase\expandafter{\romannumeral1})(\%) &AMC(\uppercase\expandafter{\romannumeral1})(\%)  & ASR(\uppercase\expandafter{\romannumeral2})(\%) &AMC(\uppercase\expandafter{\romannumeral2})(\%) \\   \hline
			
			\multirow{5}{*}{DDNE \cite{2018Deep}} 
			&DDNE &\textbf{ 100(-)} & \textbf{0.02}   & \textbf{99.83(-)}  & \textbf{99.37}  \\ 
			&DynAE &94.66$\left(\downarrow5.34\right)$&	2.53 &	88.36$\left(\downarrow11.49\right)$	&98.08 \\
			&DynRNN  & 92.21 $\left(\downarrow7.79\right)$& 6.11   & 97.65$\left(\downarrow2.18\right)$   & 91.06 \\ 
			&DynAERNN  &26.91$\left(\downarrow73.09\right)$ & 28.90  & 71.32$\left(\downarrow28.56\right)$   & 79.39 \\  	 	 	 
			&E-LSTM-D  & 43.6$\left(\downarrow56.40\right)$ &	24.86  & 	67.31$\left(\downarrow32.58\right)$ 	&82.16\\  \hline
			
			\multirow{5}{*}{DynAE \cite{2019dyngraph2vec}}
			&DDNE & 99.15$\left(\downarrow0.18\right)$  &	1.47 	&99.08$\left(\uparrow3.55\right)$ &97.42 \\
			&DynAE & \textbf{99.96(-)} &	\textbf{0.31}& 	\textbf{95.68(-)}& \textbf{97.62} \\ 
			&DynRNN & 91.7$\left(\downarrow8.26\right)$ &	8.46 	&96.56$\left(\uparrow0.92\right)$ &	87.88  \\ 
			&DynAERNN  & 26.82$\left(\downarrow73.17\right)$ &23.35& 	47.07$\left(\downarrow50.80\right)$& 	72.25 \\ 
			&E-LSTM-D  & 25.49$\left(\downarrow74.50\right)$& 	28.53 &	52.31$\left(\downarrow45.33\right)$ &	75.76\\  \hline
			
			\multirow{5}{*}{DynRNN \cite{2019dyngraph2vec}}
			&DDNE&97.80$\left(\uparrow0.63\right)$ &2.48 &	98.70$\left(\uparrow0.36\right)$ &	96.94 \\  
			&DynAE  & 95$\left(\downarrow2.25\right)$ &	4.13 &	74.01$\left(\downarrow24.75\right)$& 95.29 \\ 
			&DynRNN  & \textbf{97.19(-)} &	\textbf{2.80} & \textbf{98.35(-)} &	\textbf{89.92}  \\ 
			&DynAERNN &41.93$\left(\downarrow56.86\right)$&22.46 	&55.04$\left(\downarrow44.04\right)$	&70.90 \\ 
			&E-LSTM-D  & 23.98$\left(\downarrow75.33\right)$ &30.80 &60.07$\left(\downarrow38.92\right)$ & 75.99
			\\  \hline
			
			\multirow{5}{*}{DynAERNN \cite{2019dyngraph2vec}}
			&DDNE& 84.97$\left(\uparrow34.02\right)$ &	8.80 &	97.10$\left(\uparrow33.78\right)$ &	97.55 \\  
			&DynAE  & 36.16$\left(\downarrow42.97\right)$ &	17.90 &	82.07$\left(\uparrow13.08\right)$& 	95.13 \\ 
			&DynRNN  & 53.59$\left(\downarrow15.47\right)$ &	18.67 &	95.02$\left(\uparrow30.92\right)$ &	86.77 \\ 
			&DynAERNN &\textbf{63.40(-)} &\textbf{13.40} 	&\textbf{72.58(-)}	&\textbf{72.80} \\ 
			&E-LSTM-D  & 14.62$\left(\downarrow72.94\right)$& 	30.35&	62.03$\left(\downarrow14.54\right)$& 	80.14 
			\\  \hline

			\multirow{5}{*}{E-LSTM-D \cite{0E}}
			&DDNE  &95.43$\left(\uparrow27.26\right)$ 	&3.72& 	96.96$\left(\uparrow6.13\right)$& 	98.03 \\ 
			&DynAE  & 73.07$\left(\downarrow2.56\right)$ &9.41	&85.91$\left(\downarrow5.97\right)$ &97.76  \\ 
			&DynRNN  & 85.33$\left(\uparrow13.79\right)$ 	&5.04 &	96.97$\left(\uparrow6.14\right)$ &94.83  \\ 
			&DynAERNN  & 34.34$\left(\downarrow54.21\right)$ &	22.84& 64.97$\left(\uparrow28.89\right)$& 	78.24
			\\ 
			&E-LSTM-D  & \textbf{74.99(-)}& 	\textbf{15.18}& 	\textbf{91.36(-)} &	\textbf{89.42} \\ 
			\hline
			\bottomrule
		\end{tabular}
	\end{table*}

	\subsection{Experiment Setup}
	This section describes the settings in experiments. Considering the trade-off of attack effectiveness and concealment, $\beta$ in the objective loss Equation \ref{equ11} is set to 0.5. Since the DLP models as the attack discriminator have a good performance on DLP after 100 epochs, so the pre-trained epoch of the models is 100.
	
	To avoid the contingency of the attack, we select a total of 100 links as the target links, and each attack scenario has 50 target links. More specifically, in attack scenario \uppercase\expandafter{\romannumeral1}, prediction confidence score of each target link under clean model is larger than 0.9. These links with higher confidence scores can better verify the effectiveness of the attack. In attack scenario \uppercase\expandafter{\romannumeral2}, prediction confidence score of each target link in clean model is between 0 and 0.1.
	
	To balance the concealment and the effectiveness of the attack, we set the trigger of sequence ratio $t$ = 0.05, the poison ratio $p$ = 0.05, and the node ratio $n$ = 0.05 for attack scenario \uppercase\expandafter{\romannumeral1}, while $t$ = 0.03, $p$ = 0.05 and $n$ = 0.05 for attack scenario \uppercase\expandafter{\romannumeral2}.
	
	To observe the performance comparison of trigger injection at different timestamps, trigger injection timestamp analysis sets $t$ = 0.03, $p$ = 0.03 and $n$ = 0.03. In addition, to analyze the impact of parameter changes on the attack, parameter sensitive experiments set the fixed parameter value to 0.03.
	
	In the experiments of backdoor attacks on non-deep learning methods, we chose Deepwalk \cite{14Deepwalk} and node2vec \cite{16node2vec} models. First, the sequence obtains the node embedding by the two methods. Then the features of historical moments in the sequence are superimposed together and fed into an MLP to realize the DLP. Considering the size of the graph, we choose the embedding dimension of the Deepwalk and node2vec models to be 128.
	
	Our experimental environment consists of Intel XEON 62402.6GHz x 18C (CPU), Tesla V100 32GiB (GPU), 16GiBmemory (DDR4-RECC 2666) and Ubuntu 16.04 (OS).
	
	\subsection{Overall Performance of Backdoor Attacks}
	To verify the effectiveness of Dyn-Backdoor compared with other attack methods, we conduct attack experiments in both scenarios. Specifically, attack scenario \uppercase\expandafter{\romannumeral1} indicates that existence state of link is predicted to be non-existent. Attack scenario \uppercase\expandafter{\romannumeral2} indicates that non-existent state of link is predicted to exist. 
	
	Fig. \ref{fig:1_0} shows the experimental results under attack scenario \uppercase\expandafter{\romannumeral1}. Dyn-Backdoor has the best performance among five attack methods in terms of ASR, AMC and AUC, except for the result of the DNC dataset on DynRNN. There are three possible reasons for this phenomenon. First, the competition between the trigger generator and the attack discriminator optimizes the initial-triggers. Then, the gradient information from the attack discriminator extracts the important subgraph structure of the initial-triggers to form the triggers. In addition, Dyn-Backdoor has a dynamically evolving trigger in the training stage, that is, re-optimize the trigger at a certain interval of training iterations.
	
	The AUC obtained by the backdoored model under the clean testing data is similar to the clean model. We believe that Dyn-Backdoor is aimed at a target link, and the disturbance caused by the entire graph is imperceptible. As shown in Fig. \ref{fig:1_0} (d), the ASR of Dyn-Backdoor is 45.50\% on DynRNN, while the ASR of GB is 97.00\%. The possible reason is that Dyn-Backdoor is difficult to find effective triggers by a generative way for the DNC dataset, which is the sparse and has more nodes than other datasets.
	
	Fig. \ref{fig:0_1} shows the experimental results under attack scenario \uppercase\expandafter{\romannumeral2}. It is worth noting that the trigger size of attack scenario \uppercase\expandafter{\romannumeral2} is smaller than attack scenario \uppercase\expandafter{\romannumeral1}, but the attack effect becomes better in attack scenario \uppercase\expandafter{\romannumeral2} in terms of ASR, e.g., the ASR of DDNE reaches 100\% in Fig. \ref{fig:0_1}(a) and the ASR of E-LSTM-D reaches 100\% in Fig. \ref{fig:0_1}(c). There are two possible reasons for this phenomenon. First, there is no interference from redundant neighbors between two nodes, it is easier to establish a connection between the two nodes through a trigger. Second, these models pay more attention to the existence state of links when implementing link prediction tasks.

	\subsection{Attack Transferability}
	Since in most practical situations, the attacker may not grasp the detail of the target DLP model in prior, it is more practical to conduct a black-box setting attack, i.e., without any structure or parameter information of the DLP model. To verify the effect of the Dyn-Backdoor under black-box setting, we adopt one DLP model as the attack discriminator, and transfer the generated trigger to backdoor other DLP as the target models, named as transferable attack. Table \ref{tab:trans} shows the transferability attack results on Radoslaw.
	
	In attack scenario \uppercase\expandafter{\romannumeral1}, we find that Dyn-Backdoor's attack effect is significant against DDNE model, e.g. the ASR of attacking DDNE reaches 99.15\%, using DynAE as the attack discriminator in Table \ref{tab:trans}. When attacking the DDNE model, ASR can reach more than 80\%. This shows that DDNE has a strong ability to capture the network structure, and its robustness needs to be further strengthened. Dyn-Backdoor fails to achieve satisfactory results on DynAERNN and E-LSTM-D models, e.g. the ASR of attacking DynARRNN reaches 63\%, using DDNE as the attack discriminator and the ASR of attacking E-LSTM-D reaches 58.89\%, using DynRNN as the attack discriminator. We believe that they have more neural network layers than other models, which means that the information generated by the guided trigger is dispersed into more neural network layers so that the generative approach is difficult to generate effective triggers.

	\begin{figure*}[htb]
		\centering
		\includegraphics[width=14cm, height=4cm]{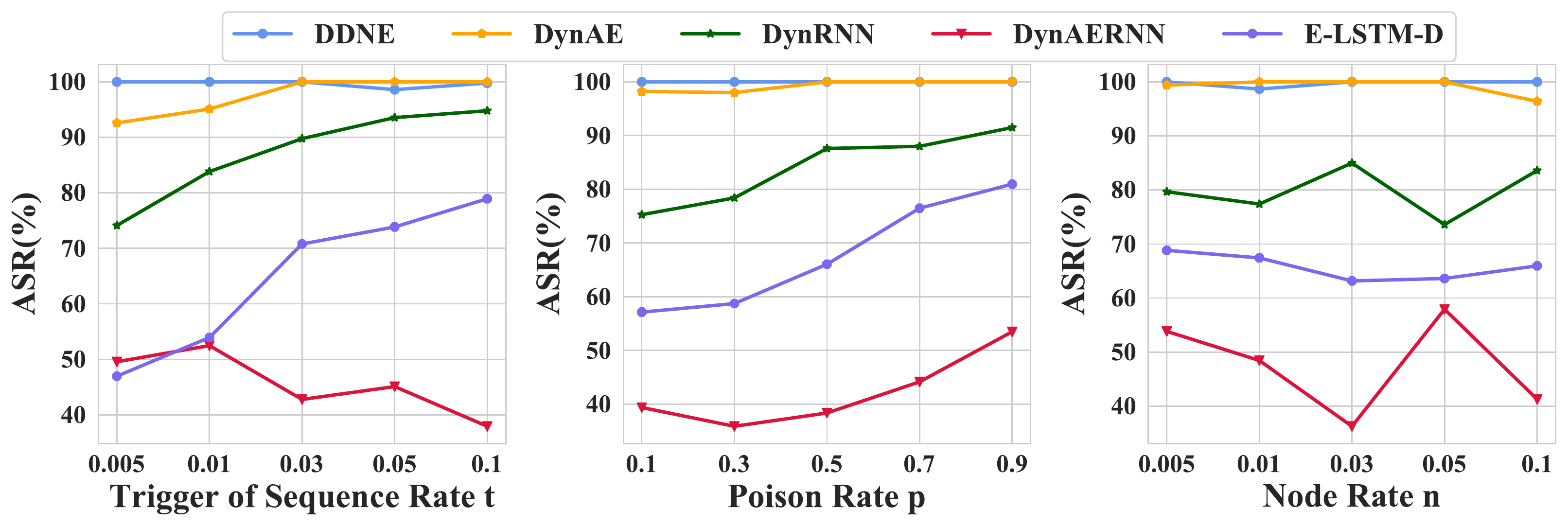}
		\caption{Parameter sensitivity analysis of Dyn-Backdoor of $t$, $p$ and $n$ on Fb-forum dataset. }
		\label{fig:params}
	\end{figure*}

	In attack scenario \uppercase\expandafter{\romannumeral2}, Dyn-backdoor can maintain good performance as well by achieving over 99\% ASR against several models, e.g. DDNE, DynRNN and DynARRNN. We find that the attack effects on the DDNE and DynAE models are very similar, e.g. the ASR of DDNE achieves 99.15\% and the ASR of DynAE achieves 99.96\%  in Table \ref{tab:trans} on DynAE as attack discriminator. The possible reason is that the encoder and decoder structures used by the two models are relatively similar. Although the models capture feature information in different ways, most DLP methods can maintain good performance. The attacker chooses a good-performance DLP model as the attack discriminator, and can perform Dyn-Backdoor under the black-box setting.

	\subsection{Parameter Sensitivity}
	The performance of Dyn-Backdoor will be mainly affected by three sensitive parameters: 1) the trigger of sequence ratio $t$; 2) the poison rate $p$; and 3) the node rate $n$. In the following, we will investigate their influences on the Dyn-Backdoor performance. According to the above experiments, the attack is more difficult to implement in attack scenario \uppercase\expandafter{\romannumeral1}, so we conduct parameter sensitivity analysis in attack scenario \uppercase\expandafter{\romannumeral1}. Fig. \ref{fig:params} shows the parameter sensitivity experiment on Fb-forum dataset. The Appendix B shows the parameter experiment results of other datasets.
	
	When exploring the influence of $t$, Fig. \ref{fig:params} can be observed that as the proportion of sequence triggers increases, except for the result on DynAERNN, the ASR will gradually increase. Intuitively, the larger the trigger size, the easier for the model to capture its structural features. However, when the $t$ increases on the DynAERNN model, the ASR decreases. It may be that the increase of the $t$ causes the original logic in the data to be destroyed, which makes the model unable to learn the characteristics of the trigger.
	
	Fig. \ref{fig:params}  shows that with the increase of the $p$, the corresponding ASR will gradually increase. When exploring the influence of $n$, we can observe that the influence of $n$ and $p$ on the model trend are very similar. This also shows that nodes with a similar structure to the nodes of target link will affect the judgment of the target link in the model learning process. In other words, we can perform backdoor attacks on the model through other similar nodes, which can also affect the target link.

	\section{Conclusion}
	This is the first work focuses on backdoor attack on DLP. To address the problem, we propose a backdoor attack framework on DLP, named Dyn-Backdoor, by adopting GAN and gradient exploration to form a trigger. Extensive experiments show the effectiveness of Dyn-Backdoor on DLP. The dynamic model will be left behind by training data with the trigger. Attacker can launch an attack by calling the trigger. 
	
	However, Dyn-Backdoor is still challenged in several ways. When dealing with large scale datasets, the convergence is much slower than small ones to generate triggers.
	Dyn-Backdoor requires the target model to provide feedback information, and the backdoor attacks on the black-box setting is also a possible future research direction.

	
	%


	\bibliographystyle{IEEEtran}      
	\bibliography{Reference}                        

	
	\ifCLASSOPTIONcaptionsoff
	\newpage
	\fi

	
	
	%
	
	%
	
	\begin{IEEEbiography}[{\includegraphics[width=1in,height=1.25in,clip,keepaspectratio]{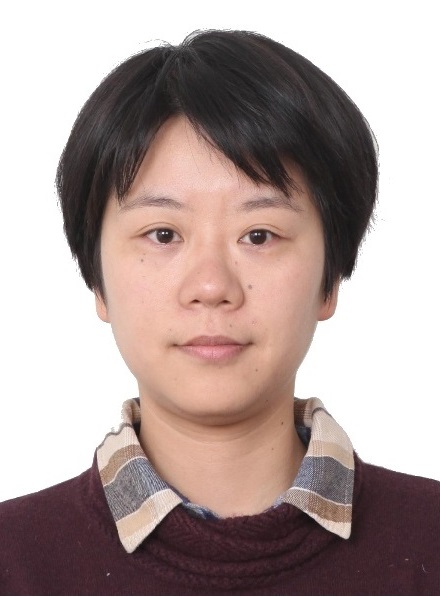}}]{Jinyin Chen}
		received BS and PhD degrees from Zhejiang University of Technology, Hangzhou,
		China, in 2004 and 2009, respectively. She studied evolutionary computing in Ashikaga Institute of Technology, Japan in 2005 and 2006.
		
		She is currently a Professor with the Zhejiang University of Technology, Hangzhou, China. Her research interests include artificial intelligence security, graph data mining and evolutionary computing.
	\end{IEEEbiography}

	\begin{IEEEbiography}[{\includegraphics[width=1in,height=1.25in,clip,keepaspectratio]{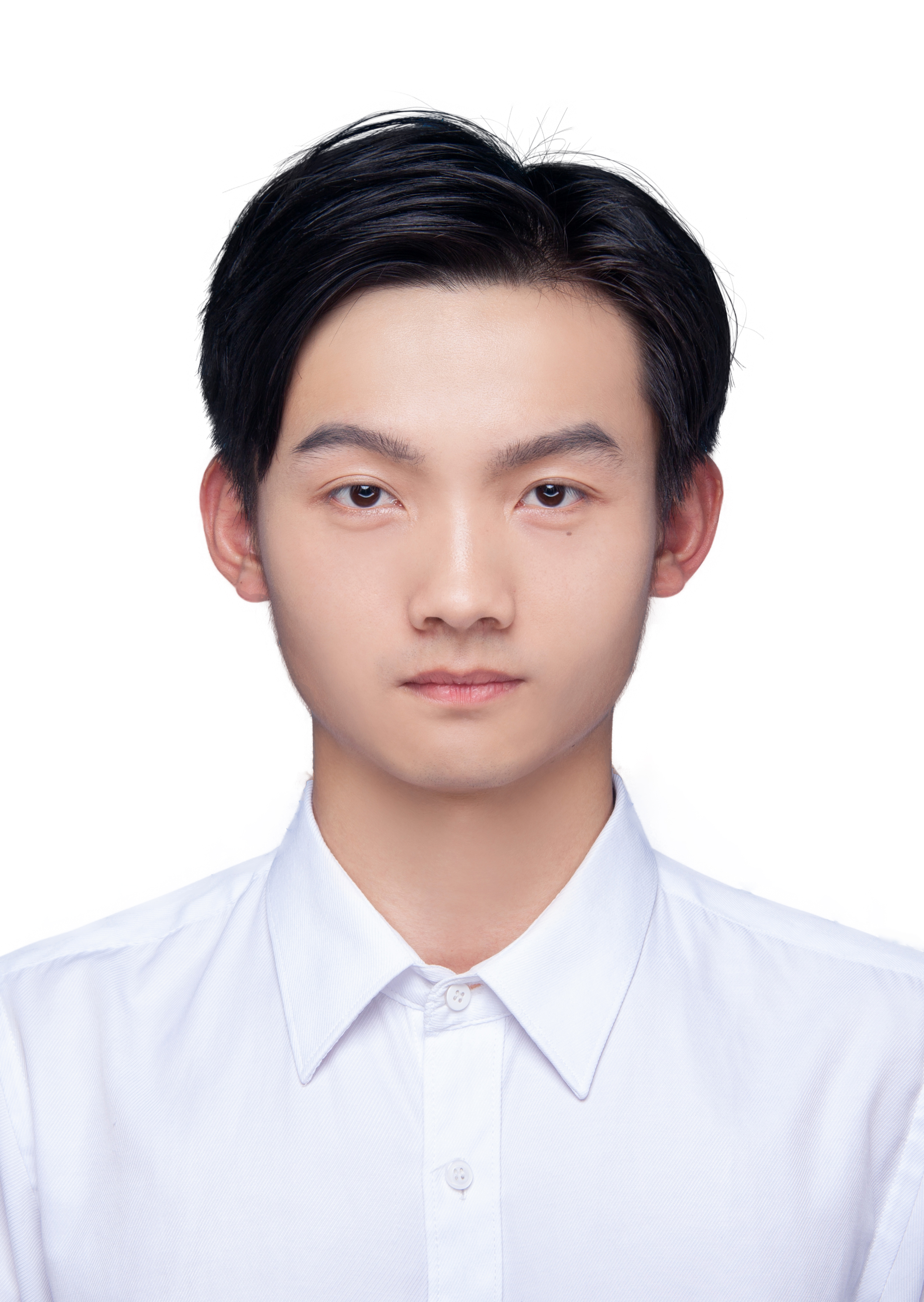}}]{Haiyang Xiong} is currently pursuing the masters degree with the college of Information engineering, Zhejiang University of Technology. 
		
		His research interests include graph data mining and applications, and artificial intelligence.
	\end{IEEEbiography}
	
	\begin{IEEEbiography}[{\includegraphics[width=1.0in,height=1.25in,clip,keepaspectratio]{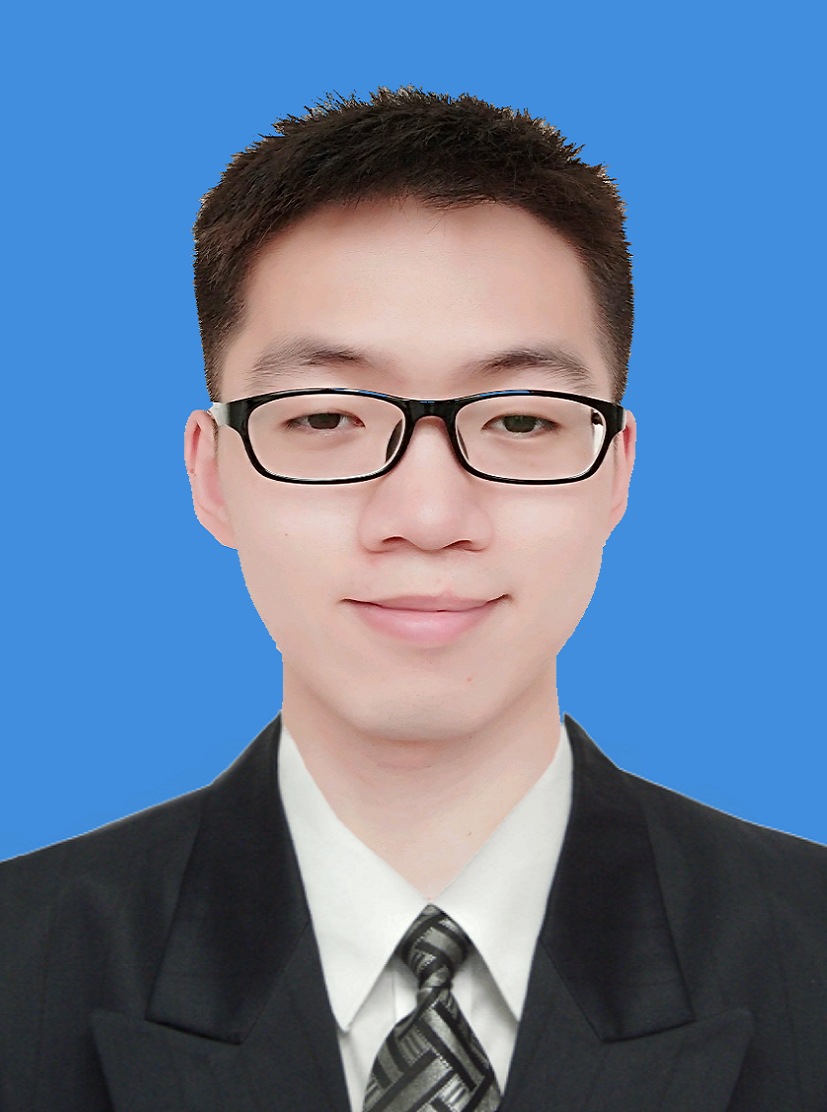}}]{Haibin Zheng}
		is a PhD student at the college of Information Engineering, Zhejiang University of Technology.
		He received his bachelor degree from Zhejiang University of Technology in 2017.
		His research interests include deep learning, artificial intelligence, and adversarial attack and defense.
	\end{IEEEbiography}
	
	\begin{IEEEbiography}[{\includegraphics[width=1in,height=1.25in,clip,keepaspectratio]{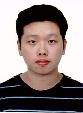}}]{Jian Zhang}
		received BS degree in automation from Zhejiang University of Technology, Hangzhou, China, in 2017.
		
		He is currently working toward the PhD degree in the College of Information Engineering, Zhejiang University of Technology, Hangzhou, China.
	\end{IEEEbiography}
	
	\begin{IEEEbiography}[{\includegraphics[width=1in,height=1.25in,clip,keepaspectratio]{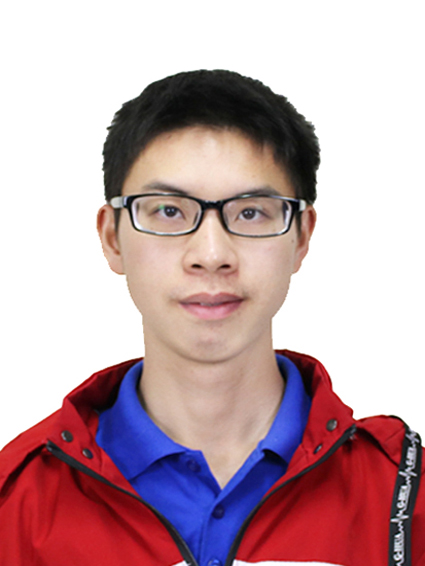}}]{Guodong Jiang}
		is currently pursuing the masters degree with the college of Information engineering, Zhejiang University of Technology. 
		
		His research interests include graph data mining and applications, and artificial intelligence.
	\end{IEEEbiography}
	
	\begin{IEEEbiography}[{\includegraphics[width=1.0in,height=1.25in,clip,keepaspectratio]{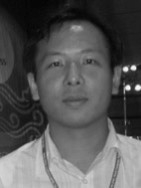}}]{Yi Liu}
		received the Ph.D. degree in control theory and engineering from Zhejiang University, Hangzhou, China, in 2009. He was an Associate Professor with the Institute of Process Equipment and Control Engineering, Zhejiang University of Technology from 2011 to 2020. He was a Postdoctoral Researcher with the Department of Chemical Engineering, Chung-Yuan Christian University from February 2012 to June 2013. Since December 2020, he has been a Full Professor with Zhejiang University of Technology, Hangzhou, China. He has published over 50 research papers at IEEE Transactions and international journals.
		
		His research interests include data intelligence with applications to modeling, control, and optimization of industrial processes.
	\end{IEEEbiography}


	
	

\end{document}